\documentclass[letterpaper, 10 pt, conference]{ieeeconf}  
\IEEEoverridecommandlockouts                              
\usepackage{hyperref}

\usepackage[utf8]{inputenc}
\usepackage[T1]{fontenc}

\usepackage{aecompl}

\usepackage{graphics} 
\usepackage{subfigure}
\usepackage{epsfig} 

\usepackage{amsthm,amsmath,amssymb}
\usepackage{mathrsfs}
\usepackage{algorithm} 
\usepackage{algorithmicx, algpseudocode}
\usepackage{multirow}
\usepackage{multicol}
\usepackage{makecell}
\usepackage{booktabs}
\usepackage{color}
\usepackage[skip=0.8pt]{caption}
\usepackage{cite}

\title{\LARGE \bf
Offline Goal-Conditioned Reinforcement Learning for Safety-Critical Tasks with Recovery Policy
}

\author{Chenyang Cao, Zichen Yan, Renhao Lu, Junbo Tan$^{*}$ and Xueqian Wang$^{*}$
\thanks{$^{}$All of the authors are with the Center for Intelligent Control and Telescience, Tsinghua Shenzhen International Graduate School, 
        Shenzhen, China.
        \texttt{ccy22@mails.tsinghua.edu.cn, tjblql@sz.tsinghua.edu.cn, wang.xq@sz.tsinghua.edu.cn}}%
\thanks{$^{*}$Corresponding author: \textit{Junbo Tan} and \textit{Xueqian Wang}}
}
\begin{document}

\maketitle
\thispagestyle{empty}
\pagestyle{empty}

\begin{abstract}

Offline goal-conditioned reinforcement learning (GCRL) aims at solving goal-reaching tasks with sparse rewards from an offline dataset. While prior work has demonstrated various approaches for agents to learn near-optimal policies, these methods encounter limitations when dealing with diverse constraints in complex environments, such as safety constraints. Some of these approaches prioritize goal attainment without considering safety, while others excessively focus on safety at the expense of training efficiency. In this paper, we study the problem of constrained offline GCRL and propose a new method called Recovery-based Supervised Learning (RbSL) to accomplish safety-critical tasks with various goals. To evaluate the method performance, we build a benchmark based on the robot-fetching environment with a randomly positioned obstacle and use expert or random policies to generate an offline dataset. We compare RbSL with three offline GCRL algorithms and one offline safe RL algorithm. As a result, our method outperforms the existing state-of-the-art methods to a large extent. Furthermore, we validate the practicality and effectiveness of RbSL by deploying it on a real Panda manipulator. Code is available at https://github.com/Sunlighted/RbSL.git.

\end{abstract}

\section{INTRODUCTION}


Reinforcement Learning (RL) enables robots to learn different skills automatically and has achieved great success in many robotic tasks such as navigation, manipulation, locomotion, etc\cite{smith2022legged, wilson2020learning, fu2023deep, abeyruwan2023sim2real, kumar2023graph}. Goal-conditioned reinforcement learning is a significant branch of RL, which aims at learning skills to reach distinct goals by adding goal information into the state\cite{rauber2018hindsight, andrychowicz2017hindsight, ghosh2020learning}. Recently, offline goal-conditioned reinforcement learning has garnered attention among researchers. It utilizes a supervised learning approach to train general-purpose policies from an offline dataset\cite{yang2021rethinking}. Thus, robots can avoid incurring real-world training costs without interacting with the environment. 
However, actions generated by such methods are unconstrained, which potentially results in dangerous behaviors, such as collision with surrounding objects, or even causing damage to the robot itself. 
For example, a robot attempts to manipulate an object to a specified location while a glass bottle blocks the path. If it ignores the obstacle, it may break the glass and cause a mess. Therefore, it is imperative for the agent to acquire a safe policy that prevents violations of constraints.


Prior works in offline GCRL\cite{chebotar2021actionable, yang2021rethinking, ma2022offline, hejna2023distance, fang2023generalization, mezghani2023learning} focus on reaching the goals along the shortest path by interacting with an offline dataset. Since the agent cannot further explore the environment to get the desired goal, imitation learning is used to solve the sparse reward and out-of-distribution (OOD) problems\cite{yang2021rethinking}. However, it probably ignores security during policy training due to violations of constraints\cite{brunke2022safe}. 
In our experiments, constraint violations tend to impede task completion, so planning a path to bypass the regions beyond the constraints will lead to better performance. To achieve this goal, we propose enhancing offline goal-conditioned reinforcement learning methods by incorporating mechanisms for constraint management.


By defining a cost function, the RL problem with constraints is called a constrained Markov decision process (CMDP)\cite{altman1995constrained}. Many prior works integrate constraint satisfaction into the optimization problem and set a Lagrange multiplier to formulate an unconstrained problem\cite{chow2017risk}. In the offline setting, several recent works try to bridge the gap between offline RL and safe RL, such as penalizing unseen actions\cite{xu2022constraints} or constructing an optimization problem with a stationary distribution correction technique\cite{lee2021coptidice}. 
The above-mentioned methods are limited in two aspects. Firstly, the training efficiency is hindered by the conflict between seeking a more optimal policy and adhering to cost constraints. Secondly, there exist estimation errors in the value functions that prevent effective cost reduction below the constraint threshold.


In this paper, we propose Recovery-based Supervised Learning (RbSL) to overcome the above shortcomings. It can be divided into two policies: the goal-conditioned policy and the recovery policy. Utilizing the advantages of imitation learning, we first train an offline policy with hindsight relabeling\cite{andrychowicz2017hindsight} and OOD action detection\cite{chebotar2021actionable} to learn goal-reaching tasks. Next, we propose a recovery policy to keep the running trajectory away from the unsafe area\cite{thananjeyan2021recovery}. Such structure design helps balance the targets of goal attainment and constraint satisfaction. 


In addition, the quality of the dataset also has a significant effect on training results, which is rarely discussed in prior works. Hence, in order to achieve better performance of the recovery policy, we filter out unsafe trajectories from the dataset and keep successful ones to guide safe actions at the boundary of constraints. Besides, we undertake a cost-value relabeling strategy to improve the training efficiency of the recovery policy.  

The main contributions are summarized as follows: 

\begin{itemize}

\item Supervised learning and reinforcement learning are integrated to solve the offline problem. Besides, the OOD action detection technique is proposed to further enhance the policy performance.
\item Separate policies are used to ensure safety for the task and constraints without a substantial reduction in the success rate.
\item The recovery dataset is reshaped to improve the safety of the recovery policy.

\end{itemize}

\section{RELATED WORK}

\subsection{Goal-Conditioned Reinforcement Learning}



Goal-conditioned reinforcement learning tries to solve problems with sparse rewards and multiple goals, which has been studied in many works \cite{andrychowicz2017hindsight, rauber2018hindsight, lin2020invariant, ghosh2020learning, dengler2022learning}. Hindsight experience relabeling (HER)\cite{andrychowicz2017hindsight} is a method that relabels the failure trajectories to successful experiences to maximize training efficiency. HPG\cite{rauber2018hindsight} combines HER with policy gradient to achieve a stable estimation of the goal-conditioned value function. GCSL\cite{ghosh2020learning} uses HER and imitation learning to learn an optimal policy. 

Many works extend the above-mentioned methods to an offline setting, where agents do not interact with the environment. WGCSL\cite{yang2021rethinking} extends GCSL to the offline setting by using weighted supervised learning. GoFAR\cite{ma2022offline} utilizes value iteration over a dual form of GCRL. Other methods like Actionable Model\cite{chebotar2021actionable} are based on Q-learning and attempt to address OOD errors by reducing the selection of actions outside the dataset. DQAPG\cite{yang2023swapped} utilizes the method of goal-swapping for data augmentation, which improved offline algorithms from another perspective. To solve long-horizon problem, Li et al.\cite{li2023hierarchical} propose a hierarchical trajectory-level diffusion (HDMI) to discovery subgoals and generate actions. Our work is inspired by WGCSL and concentrates on improving the ability to ensure security.

\subsection{Offline Safe Reinforcement Learning}


Safe RL algorithms are used to solve constrained optimization problems\cite{altman1995constrained, achiam2017constrained, chow2017risk, tessler2018reward, zhang2022safety, lin2023safe}. Lagrangian method\cite{chow2017risk} is the most common way to deal with constraints among these algorithms, which adds a multiplier to penalize the constraint item. However, Lagrangian-based policy faces difficulties in balancing the trade-off between safety and exploration. What's more, in the offline setting, it performs poorly in limiting the cost value especially when the expert data is not sufficient. Several prior works have explored methods for minimizing constraint violations during offline training\cite{le2019batch, xu2022constraints, lee2021coptidice, liu2023constrained}. They incorporate offline RL ideas\cite{levine2020offline, prudencio2023survey} into safe RL, such as using the stationary distribution correction estimation technique to optimize the policy with a cost upper bound\cite{lee2021coptidice}. CPQ\cite{xu2022constraints} is a value-based method that labels the OOD actions unsafe to reduce the estimation error of cost value. 


Another type of safe RL is called Recovery RL \cite{thananjeyan2021recovery}, which is composed of a task policy and a recovery policy. Its main idea is to use a pre-trained policy to shield the agent from hazardous areas. Guided by a risk Q-value, the recovery policy is designed to adapt unsafe actions to ensure safety\cite{alshiekh2018safe}. Building upon Recovery RL, our method exploits its potential in an offline setting, a domain that has received limited attention in prior research.

\section{PRELIMINARIES}



\subsection{Constrained Goal-augmented Markov Decision Process}

We propose a Constrained Goal-augmented Markov Decision Process (CGMDP) as a variant of a Goal-augmented Markov Decision Process (GMDP). A CGMDP can be expressed as a tuple $(\mathcal{S}, \mathcal{A}, \mathcal{G}, \mathcal{P}, \mathcal{R}, \gamma, \mathcal{C})$. $\mathcal{S},  \mathcal{A}, \mathcal{G}, \mathcal{P}$ denotes state, action, goal space and transition probability, respectively. $\gamma$ is a discount factor and $r(s,a,g)$ = $1[\Vert \phi(s) - g \Vert ^2_2 \leq \delta] : \mathcal{S}\times\mathcal{A}\times\mathcal{G} \rightarrow \mathcal{R} $ is a goal-conditioned binary reward related to a threshold $\delta$. $\phi$ means a state-to-goal mapping. The cost function is $c(s,a): \mathcal{S}\times\mathcal{A} \rightarrow \mathcal{C}$. The cost value $V_C^{\pi}$, cost action-value $Q_C^{\pi}$ and cost advantage functions $A_C^{\pi}$ are defined by
\begin{equation}
\begin{aligned}
     V_C^{\pi}(s,g) &:= \mathbb{E}_{\pi} \big[ 
     \sum^{\infty}_{t=0} \gamma^{t} c(s_t, a_t) | s_0 = s, g \big], \\
     Q_C^{\pi}(s,a,g) &:= \mathbb{E}_{\pi} \big[ 
     \sum^{\infty}_{t=0} \gamma^{t} c(s_t, a_t) | s_0 = s ,a_0 = a, g\big], \\
     A_C^{\pi}(s,a,g) &:= Q_C^{\pi}(s,a,g) - V_C^{\pi}(s,g).
\end{aligned}
\end{equation}

In the CGMDP setting, a policy $\pi : \mathcal{S}\times \mathcal{G} \rightarrow \mathcal{A}$ aims to maximize the expected return with constraints:
\begin{equation}
\begin{aligned}
    J(\pi) = \max_{\pi} \,\, & \mathbb{E}_{g\sim p(g),a_t \sim \pi(\cdot |s_t,g)} \big[\sum^{\infty}_{t=0} \gamma^{t} r(s_t,a_t,g)\big], \\
     & \mathrm{s.t.} \,\, J_C(\pi) \leq l,
\end{aligned}
\end{equation}
where $J_C(\pi) = \mathbb{E}_{a_t \sim \pi(\cdot |s_t,g)} \big[\sum^{\infty}_{t=0} \gamma^{t} c(s_t, a_t) \big]$ denotes the expectation of discounted cost return and $l$ is the constraint bound. This constrained optimization problem can be modified into a Lagrangian form:
\begin{equation}
     \min_{\lambda\geq 0} \max_{\pi} \ \mathbb{E}\big[ 
     \sum^{\infty}_{t=0} \gamma^{t} (r(s_t,a_t,g) - \lambda c(s_t, a_t)) + \lambda l\big],
\end{equation}
where $\lambda$ is a Lagrange multiplier.

\subsection{Offline Constrained GCRL}

In an offline setting, the training trajectory is sampled from a static dataset of logged transitions $\mathcal{D}:= \{ \tau_i\}_{i=1}^N$ in a form of $\tau_{i} = (s_0^{(i)}, a_0^{(i)}, r_0^{(i)}, c_0^{(i)}\dots; g^{(i)})$. The optimization problem of constrained GCRL can be described as:
\begin{equation} \label{cgcrl}
\begin{aligned}
    J(\pi) = & \max_{\pi} \,\, \mathbb{E}_{(s_t,s_{t+1},g)\sim \mathcal{D}, a_t\sim \pi} \big[ \sum^{\infty}_{t=0} \gamma^{t} r(s_t,a_t,g)\big], \\
     & \mathrm{s.t.} \,\, \mathbb{E}_{(s_t,s_{t+1})\sim \mathcal{D}, a_t\sim \pi} \big[ 
     \sum^{\infty}_{t=0} \gamma^{t} c(s_t, a_t) \big] \leq l.
\end{aligned}
\end{equation}

\begin{figure*}[!ht]
\centering
\includegraphics[trim=0cm 0cm 0cm 0cm,clip,width=0.9\linewidth]{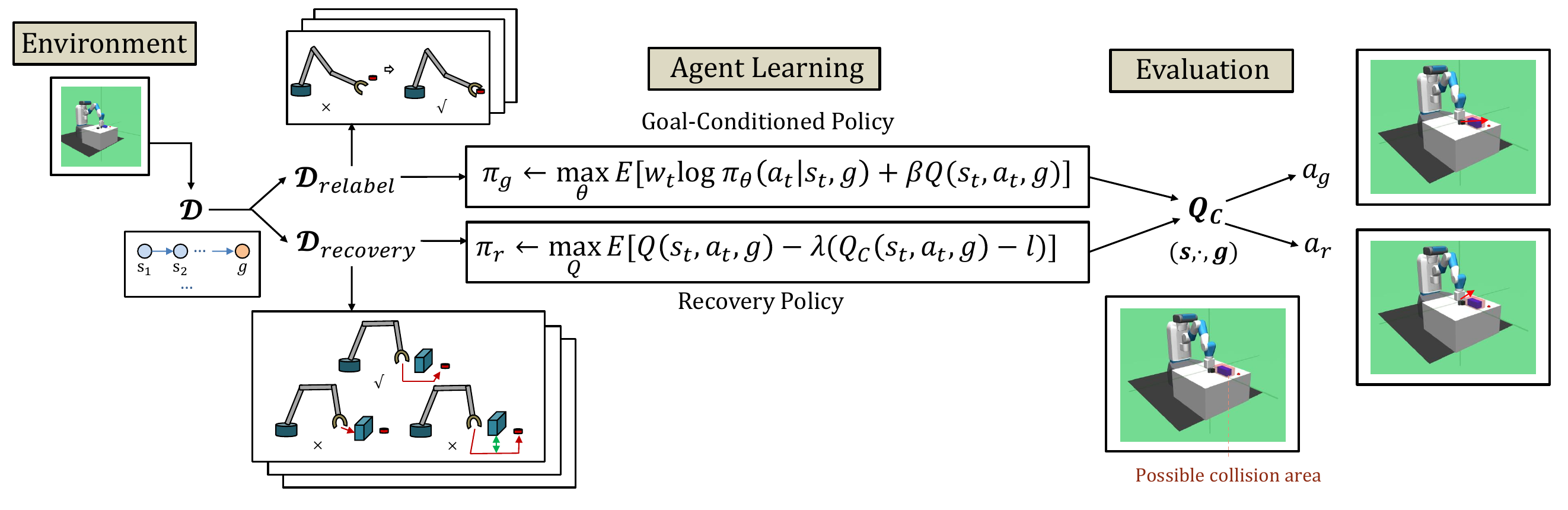}\\
\centering
\caption{An overview of Recovery-based Supervised Learning (RbSL). RbSL first samples data from the environment and processes them into two datasets. Then, the agent learns a goal-conditioned policy and a recovery policy. In evaluation, we use the cost Q-value to predict an unsafe state and decide which policy to use.}
\label{fig:overview}
\end{figure*}

\section{METHOD}



The proposed method simplifies the problem \eqref{cgcrl} into two parts. The safe goal-reaching problem is addressed through a supervised learning-based method, while constraints are managed by a recovery policy. The combination of these two policies significantly improves training efficiency.

\subsection{Overview}
In order to reach goals safely, we train a cost Q-network to predict the likelihood of violations, which determines the switching between two policies:
\begin{equation}
\begin{aligned}
       \pi(a_t|s_t,g) = \begin{cases}\pi_{\textrm{g}}(a_t|s_t,g), & Q_C(s_t,a_t,g) \leq l,
       \cr \pi_{\textrm{r}}(a_t|s_t,g), &Q_C(s_t,a_t,g) > l, \end{cases}
\end{aligned}
\end{equation}
where $\pi_{\textrm{g}}$ indicates the goal-conditioned policy and $\pi_{\textrm{r}}$ denotes the recovery policy. 
The goal-conditioned policy is primarily oriented toward reaching the goal. When it breaches the constraint, the recovery policy will try to modify the action in the direction of decreasing the cost Q-value. Our proposed method is illustrated in Figure \ref{fig:overview}. Next, in sections B and C, we will introduce more details of the two policies and discuss why they behave well in the offline setting. 

\subsection{Supervised Learning}

In offline GCRL, imitation learning methods like GCSL\cite{ghosh2020learning} mimic actions to ensure that the learned policy is consistent with the data distribution.  However, it may perform sub-optimal behavior in the offline setting as it only focuses on the success of each task at the end of the episode while neglecting optimization at each individual step. To overcome this disadvantage, prior works have tried to estimate the optimal policy by advantage-weighted regression\cite{peng2019advantage, kostrikov2021offline}. Weighted goal-conditioned supervised learning (WGCSL) can determine the shortest path in offline goal-reaching tasks. Therefore, we use it as our goal-conditioned policy.

First, we relabel the offline dataset $\mathcal{D}$ for supervised learning. Suppose that $\mathcal{D} = \{s_t,a_t,r_t,g\}, t\in [1,T]$, we relabel the goal with a future state $g' = \phi(s_{t'}), t'\geq t$, where $\phi$ is a mapping from state to goal. In an environment with constraints, we add the information of constraints as a part of the state into the dataset:
\begin{equation}
\begin{aligned}
       \mathcal{D}_{\textrm{relabel}} = \begin{cases}\{s_t,a_t,r_{t'},g'= \phi(s_{t'}), o \}, & s_T \notin o, \cr \{s_t,a_t,r_t,g,o\}, &\text{otherwise},\end{cases}
\end{aligned} \label{eq:relabel}
\end{equation}
where $o$ indicates an unsafe area. The relabeling reduces the experience that violates the constraints and increases successful trajectories to improve learning efficiency.

Next, we use a composed weight to guarantee an optimal policy in the offline setting. The policy can be trained by maximizing the following BC-based expectations\cite{torabi2018behavioral}: 
\begin{equation}
    J_{\mathrm{wgcsl}}(\pi) = \mathbb{E}_{(s_t,a_t,g')\sim \mathcal{D}_{\textrm{relabel}}} \big[w_{t,t'} \cdot \log\pi_{\theta}(a_t|s_t,g') \big] \label{eq:wgcsl},
\end{equation}
where the weight $w_{t,t'} = \gamma^{t'-t} \cdot \exp(A(s_t,a_t,g')) \cdot \epsilon(A(s_t,a_t,g'))$. The first part $\gamma^{t'-t}$ assigns smaller weight to sub-optimal trajectories. The second part $\exp(A(s_t,a_t,g')) $ restricts the learned policy to avoid OOD actions. The last part equals one if $A(s_t,a_t,\phi(s_i)) > \hat{A}$ and equals to $\epsilon$ otherwise, where $\hat{A}$ is a threshold and $\epsilon$ is a small value. It leads the policy to the highest return.

To get better performance than imitation learning on goal-reaching tasks, we consider optimizing the Q-value to improve the behavior policy. Therefore, we rewrite the policy improvement \eqref{eq:wgcsl} following TD3+BC\cite{fujimoto2021minimalist} approach:
\begin{equation}
\begin{aligned}
    J(\pi) = \ & \mathbb{E}_{(s_t,a_t,g')\sim \mathcal{D}_{\textrm{relabel}}} \big[w_{t,t'} \cdot \log\pi_{\theta}(a_t|s_t,g')
    \\ &+ \alpha' Q_g(s_t,a_t,g') \big] \label{eq:task_policy},\\
     \textrm{and\ \ }&  \alpha' = \frac{\alpha}{\frac{1}{N}\sum_{(s_i,a_i)}|Q_g(s_i,a_i,g')|},
\end{aligned}
\end{equation}
where $\alpha$ is a hyper-parameter to balance maximizing Q-value and minimizing the BC term. Since $\alpha$ is highly influenced by the scale of Q-value, we add a normalization term $\frac{1}{N}\sum_{(s_i,a_i)}|Q_g(s_i,a_i,g')|$ into it.

\subsection{Recovery Policy}

\begin{figure}[t]
\centering
\includegraphics[trim=0cm -0.5cm 0cm 0cm,clip,width=0.9\linewidth]{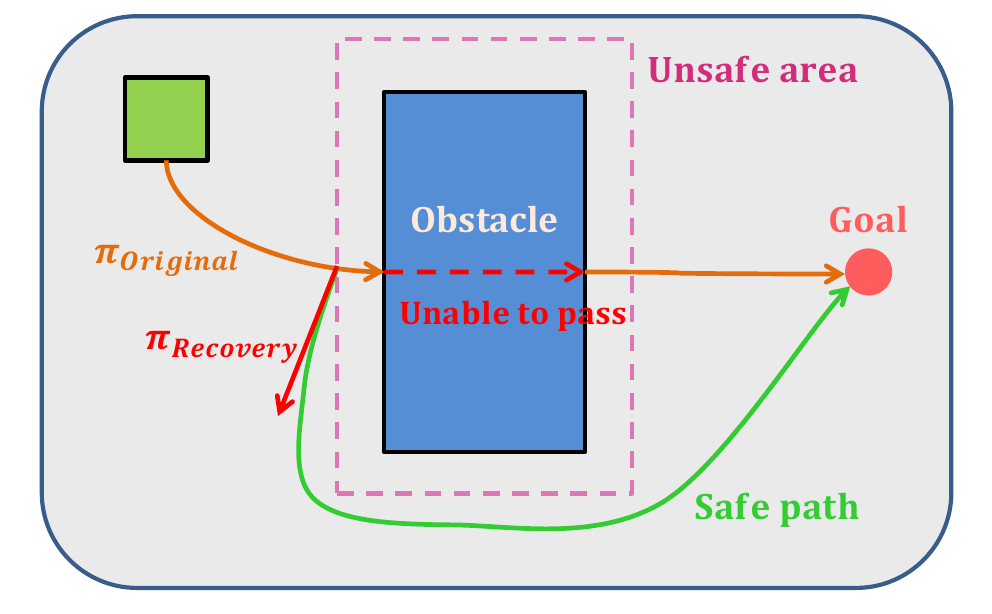}\\
\centering
\caption{Recovery policy: we illustrate the recovery policy on a robotic push task. The original policy will choose the shortest path to reach the goal while the obstacle blocks its way and results in failure. In contrast, the recovery policy will correct the action of entering into an unsafe area and plan a safe path to the goal.}
\label{fig:recovery}
\end{figure}

The recovery policy is responsible for action correction as shown in Figure \ref{fig:recovery}. We first learn a constraint Q-value $Q_C(s_t,a_t,g)$ to estimate the future probability of violation:
\begin{equation}
\begin{aligned}
    Q_C&(s_t,a_t,g) = \mathbb{E}\big[\sum^{\infty}_{t'=t} \gamma^{t'-t} c_{t'} | s_t ,a_t \big]
    \\ &=c_t + (1 - c_t)\gamma \mathbb{E}\big[ Q_C(s_{t+1}, a_{t+1}, g)| s_t, a_t\big],
\end{aligned}\label{eq:qcost}
\end{equation}
where $c_t$ indicates the cost function. Recovery RL\cite{thananjeyan2021recovery} pretrains the cost Q-value on the historical trajectories that contain constraint violations. In contrast, our method trains the cost Q-value and the goal-conditioned Q-value simultaneously as the trajectories contain information about a complete path. It is necessary for the agent to learn to reach the goal so that the recovery policy can be effectively updated. To this end, we consider policy improvement as follows:
\begin{equation}
\begin{aligned}
    \pi_r = & \mathop{\arg\min}\limits_{\pi} \mathbb{E}_{(s,g)\sim \mathcal{D}, a\sim \pi(\cdot| s,g)}\big[Q_r(s, a, g ) \\
    &- \lambda (Q_C(s, a, g ) - l)\big],
\end{aligned}\label{eq:recovery}
\end{equation}
where $\lambda$ represents a Lagrange multiplier in an optimization problem, chosen to be sufficiently large to constrain the cost Q-value below $l$. 

In the offline setting, there exists a problem of overestimating Q-values caused by OOD actions for the recovery policy. Hence, we use $\mathcal{A}'(s,g)$ to denote an unseen action set that can not be reached when the agent tries to achieve the goal $g$ from the state $s$ in the dataset $\mathcal{D}$. Next, let $\mathcal{G}'(g)$ be a set of negative goals different from the desired goal $g$. We can assume that any action $a' \sim \mathcal{A}'(s,g)$ finally leads to a negative goal $g' \sim \mathcal{G}'(g)$. Therefore, we can minimize the Q-value for these negative actions $a'$ using relabeling. We first sample a negative action $a'$ from the soft-max distribution $\frac{exp(Q_r(s,a',g))}{Z}$ where $Z$ is a normalization factor. Then we set the target Q-value to zero as $Q_{target}(s,a',g)\leftarrow 0$ so that we alleviate the high value of OOD action and get the correct recovery policy.

\subsection{Data Processing}
In the offline setting, we should collect an offline dataset $\mathcal{D}$ containing data from random policies and experts first. The task policy and recovery policy are both trained on $\mathcal{D}$. However, the recovery policy is so sensitive to random data that safe actions cannot be learned effectively. Additionally, some trajectories with zero cost return are useless for training recovery policy. Therefore, we filter the dataset $\mathcal{D}$ in two steps: 1) filter out random trajectories $\tau_i$ if $R(\tau_i) = \sum^T_{t=0}\gamma^t r_i(s_t,a_t,g) = 0$ to get an expert dataset $\mathcal{D}_e$; 2) filter out some expert trajectories $\tau_i'$ if $C(\tau_i') = \sum^T_{t=0}\gamma^t c_i(s_t, a_t) = 0$ from $\mathcal{D}_e$. Finally, we get a recovery dataset $\mathcal{D}_\mathrm{rec}$.

To train a recovery policy, we consider cost shaping to $\mathcal{D}_\mathrm{rec}$ as follows: 
\begin{equation}
\begin{aligned}
    c'(s_t, a_t) = \begin{cases}c(s_t, a_t)-1, & c(s_{t}, a_t) = 1 \ \mathrm{and}\ \\ & c(s_{t+1}, a_{t+1}) = 0,
       \cr c(s_t, a_t), &\mathrm{otherwise}. \end{cases}
\end{aligned}\label{eq:dataset}
\end{equation}

We reduce the cost by learning actions that prevent the policy from violating constraints. The cost shaping facilitates the recovery policy in learning such safe actions.

To sum up, two policies are trained on different datasets independently. During the evaluation phase, we use both policies for decision-making to ensure both the task performance and security. We illustrate our complete method in Algorithm \ref{alg:main}. 

\begin{algorithm}[t]
\caption{Recovery-based Supervised Learning}
\label{alg:main}
\begin{algorithmic}[1]
\Require $\mathcal{D}_\mathrm{offline}$, number of episodes $N$, number of episodes $N_r$ for recovery training, a Lagrangian multiplier $\lambda$
\State Randomly initialize policy $\pi_g$,$\pi_r$, and value function $Q_g$, $Q_r$, $Q_C$.
\State Filter the recovery dataset and get $\mathcal{D}_{\mathrm{goal}} \leftarrow \mathcal{D}_\mathrm{offline}$, $\mathcal{D}_{\mathrm{recovery}} \leftarrow \mathcal{D}_\mathrm{filter}$
\For{$i \in \{1,\ldots N\}$}
    \State Sample a minibatch $\{(s_t^i,a_t^i,g^i,s_{t+1}^i,r_t^i)\}_{i=1}^M$ from the offline dataset $\mathcal{D}_\mathrm{goal} $
    \State Relabel the minibatch by using Eq.\ref{eq:relabel}: $\{(s_t^i,a_t^i,\phi(s_j^i),s_{t+1}^i,r_t^i(\cdot,\cdot,\phi(s_j^i)))\}_{i=1}^M, j\geq t$
    \State Update $Q_g$ and estimate $A(s_t^i,a_t^i,\phi(s_j^i))$
    \State Get $K$ percentile $\hat{A}$ and update policy $\pi_g$ by Eq.\ref{eq:task_policy}
\EndFor
\For{$i \in \{1,\ldots N_r\}$}
    \State Sample a minibatch $\{(s_t^i,a_t^i,g^i,s_{t+1}^i,r_t^i,c_t^i)\}_{i=1}^M$ from the offline dataset $\mathcal{D}_\mathrm{recovery} $
    \State Relabel dataset using Eq.\ref{eq:relabel} and do cost shaping Eq.\ref{eq:dataset}
    \State Update $Q_r$ by TD-error with negative action penalty
    \State Update $Q_C$ by Eq.\ref{eq:qcost}
    \State Update recovery policy $\pi_r$ by Eq.\ref{eq:recovery}
\EndFor
\end{algorithmic}
\end{algorithm}



\begin{table*}[t]
  \centering
  \vspace{-2mm}
    \caption{Main Result (averaged over 5 seeds)}
    \resizebox{\linewidth}{!}{\begin{tabular}{cccccccccccc}
    \toprule
    \multirow{2}{*}{\centering\textbf{Task}} & \multirow{2}{*}{\textbf{\makecell{Mixture of dataset \\(expert--random)}}} & \multicolumn{2}{c}{\textbf{AM-lag}} & \multicolumn{2}{c}{\textbf{GCSL}} & \multicolumn{2}{c}{\textbf{WGCSL}}  & \multicolumn{2}{c}{\textbf{GoFAR}}  & \multicolumn{2}{c}{\textbf{RbSL(Ours)}}\\
    \cmidrule(lr){3-4} \cmidrule(lr){5-6} \cmidrule(lr){7-8} \cmidrule(lr){9-10} \cmidrule(lr){11-12}
    ~ & ~ & {\textbf{\makecell{Success \\ rate}}} & {\textbf{\makecell{Cost \\ return}}} & {\textbf{\makecell{Success \\ rate}}} & {\textbf{\makecell{Cost \\ return}}} & {\textbf{\makecell{Success \\ rate}}} & {\textbf{\makecell{Cost \\ return}}} & {\textbf{\makecell{Success \\ rate}}} & {\textbf{\makecell{Cost \\ return}}} & {\textbf{\makecell{Success \\ rate}}} & {\textbf{\makecell{Cost \\ return}}}\\
    \midrule
    \multirow{2}{*}{ReachObstacle} & 0.1-0.9 & \textbf{1}& \textbf{0.016}{\tiny$\pm$0.000} & $0.997$ & $3.15${\tiny$\pm$0.036} & \textbf{1}& $1.646${\tiny$\pm$0.004}& \textbf{1}& $1.886${\tiny$\pm$0.000}& \textbf{1}& 0.123{\tiny$\pm$0.000}\\
    ~ & 0-1 & $0.983$& $0.217${\tiny$\pm$0.071}& $0.773$ & $17.65${\tiny$\pm$1.144} & $0.99$& $6.326${\tiny$\pm$1.971} & $0.873$ & $7.996${\tiny$\pm$0.684}& $\textbf{1}$ & $\textbf{0.21}${\tiny$\pm$0.000}\\
    \midrule
    \multirow{4}{*}{PickAndPlaceObstacle} & 1-0 & $0.58$ & $1.223${\tiny$\pm$0.521}& $0.719$& $6.336${\tiny$\pm$0.859}& 0.916& $4.003${\tiny$\pm$1.21}& $0.916$& $2.966${\tiny$\pm$0.089}& \textbf{0.946} & \textbf{0.87}{\tiny$\pm$0.276} \\
    ~ & 0.5-0.5 & $0.447$& $1.902${\tiny$\pm$0.391} & $0.68$& $7.65${\tiny$\pm$1.266}& $0.86$& $4.253${\tiny$\pm$0.841}& $0.91$& $4.53${\tiny$\pm$1.353}& \textbf{0.936} & \textbf{1.41}{\tiny$\pm$0.083} \\
    ~ & 0.2-0.8 & $0.337$& $2.654${\tiny$\pm$1.833} & $0.623$ & $7.816${\tiny$\pm$0.592} & $0.81$& $5.28${\tiny$\pm$0.053}& $0.836$& $4.47${\tiny$\pm$0.405}& \textbf{0.87}& \textbf{2.14}{\tiny$\pm$0.018}\\
    ~ & 0.1-0.9 & $0.206$& $2.947${\tiny$\pm$4.172} & $0.553$ & $9.236${\tiny$\pm$3.295} & $0.733$& $9.11${\tiny$\pm$1.411}& $0.793$& $6.336${\tiny$\pm$0.341}& \textbf{0.843} & \textbf{2.516}{\tiny$\pm$0.75} \\
    \midrule
    \multirow{4}{*}{PushObstacle} & 1-0 & $0.639$& $2.473${\tiny$\pm$0.363}& $0.71$& $4.353${\tiny$\pm$0.285}& $0.813$& $3.74${\tiny$\pm$1.162} & $0.723$ & $4.026${\tiny$\pm$1.843}& \textbf{0.896}& \textbf{0.566}{\tiny$\pm$0.009}\\
    ~ & 0.5-0.5 & $0.57$& $3.2${\tiny$\pm$1.786}& $0.667$& $5.233${\tiny$\pm$2.437} & $0.756$ & $4.633${\tiny$\pm$0.991} & $0.553$ & $3.96${\tiny$\pm$0.204} & \textbf{0.883}& \textbf{0.553}{\tiny$\pm$0.015}\\
    ~ & 0.2-0.8 & $0.413$& $4.766${\tiny$\pm$1.103}& $0.523$& $6.066${\tiny$\pm$1.703} & $0.646$ & $4.71${\tiny$\pm$1.251}& $0.41$ & $5.23${\tiny$\pm$0.504}& \textbf{0.667} & \textbf{2.113}{\tiny$\pm$0.464}\\
    ~ & 0.1-0.9 & $0.083$& \textbf{1.65}{\tiny$\pm$2.63}& $0.393$& $5.4${\tiny$\pm$0.557}& $0.526$& $4.31${\tiny$\pm$0.275}& $0.343$& $5.64${\tiny$\pm$2.39}& \textbf{0.603} & 2.43{\tiny$\pm$0.569}\\
    \midrule
    \multirow{4}{*}{SlideObstacle} & 1-0 & $0.336$ & $2.483${\tiny$\pm$0.135} & $0.313$& $3.436${\tiny$\pm$0.7}& $0.33$ & $3.02${\tiny$\pm$0.326} & $0.523$& $3.239${\tiny$\pm$0.128}& \textbf{0.556} & \textbf{0.356}{\tiny$\pm$0.012} \\
    ~ & 0.5-0.5 & $0.153$& $2.81${\tiny$\pm$0.948} & $0.28$& $4.749${\tiny$\pm$1.11}& $0.313$ & $3.25${\tiny$\pm$0.268}& $\textbf{0.516}$& $3.023${\tiny$\pm$0.042}& 0.459 & \textbf{0.77}{\tiny$\pm$0.141} \\
    ~ & 0.2-0.8 & $0.05$& $1.5${\tiny$\pm$0.551} & $0.216$& $3.953${\tiny$\pm$1.233}& $0.24$& $2.846${\tiny$\pm$0.264}& $\textbf{0.353}$ & $2.153${\tiny$\pm$0.239}& 0.286 & \textbf{1.21}{\tiny$\pm$0.058} \\
    ~ & 0.1-0.9 & $0.02$& \textbf{0.85}{\tiny$\pm$1.156} & $0.176$ & $4.373${\tiny$\pm$2.040} & $0.15$ & $2.803${\tiny$\pm$0.375} & $\textbf{0.293}$ & $2.656${\tiny$\pm$0.285}& 0.243 & 1.75{\tiny$\pm$0.299} \\
    \midrule
    \textbf{Average} & - & $0.416$ & $2.049$ & $0.545$ & $6.386$ & $0.649$ & $4.281$ & $0.646$ & $4.151$ & \textbf{0.726} & \textbf{1.216} \\
    \bottomrule
  \end{tabular}}
  \label{tab:main}
\end{table*}

\section{EXPERIMENTS}
Our experiments are carried out in Gym-Robotics environments \cite{plappert2018multi} with four classic manipulation tasks. Besides, Panda-Gym\cite{gallouedec2021panda} is used for applying RbSL to the sim-to-real experiments, as shown in Figure \ref{fig:envs}. In this section, we aim to answer the following questions: 

1) Can RbSL consistently outperform prior methods across various tasks on datasets of different quality?

2) Can RbSL reach goals with fewer constraint violations compared to other offline goal-conditioned algorithms?

3) Can RbSL successfully transition from a simulated environment to real-world applications?

Moreover, in order to verify our method's robustness, we change the mixing ratio of expert and random data in the offline dataset to test the algorithm's performance. In the rest of this section, the practical significance of the proposed method is verified. 

\subsection{Environment and Experiment Setup}
We modify the four manipulation tasks in MoJoCo \cite{todorov2012mujoco} Gym-Robotics by adding obstacles: \textit{ReachObstacle, PickAndPlaceObstacle, PushObstacle}, \textit{SlideObstacle}. These tasks require the robot or object to reach the goal as quickly as possible without collision. The best case is to find the shortest path away from the obstacle to attain the goal. 

The state space for \textit{ReachObstacle} is $\mathcal{S} \in \mathbb{R}^{19}$ consists of a 10-dim vector about basic environment information and another 9-dim vector representing obstacle position and rotation. The rest three tasks have a (25+12)-dim state by adding the object information and relative positions. The action space $\mathcal{A} \in \mathbb{R}^{4}$ contains the increments of the end-effector in the Cartesian space and control of grippers.

The cost function is defined as
\begin{equation}
\begin{aligned}
       c(s_t) = 1(s_t \notin O),
\end{aligned}
\end{equation}
where $O = \{\mathbf{o} = (x,y,z)|d(\mathbf{o}_i, \mathbf{s}_i) - L_i \leq \epsilon , i = x,y,z\}$ is a box region and $\mathbf{s}$ is the object (or grasp) coordinate. $L_i$ is the sum of object and obstacle sizes in orientation $i$. $\epsilon$ is a threshold (0.05 in experiments).

\begin{figure*}[t]
\centering
\subfigure[Reach]{
    \begin{minipage}[t]{0.2\pdfpagewidth}
        \centering
        \includegraphics[width=\linewidth]{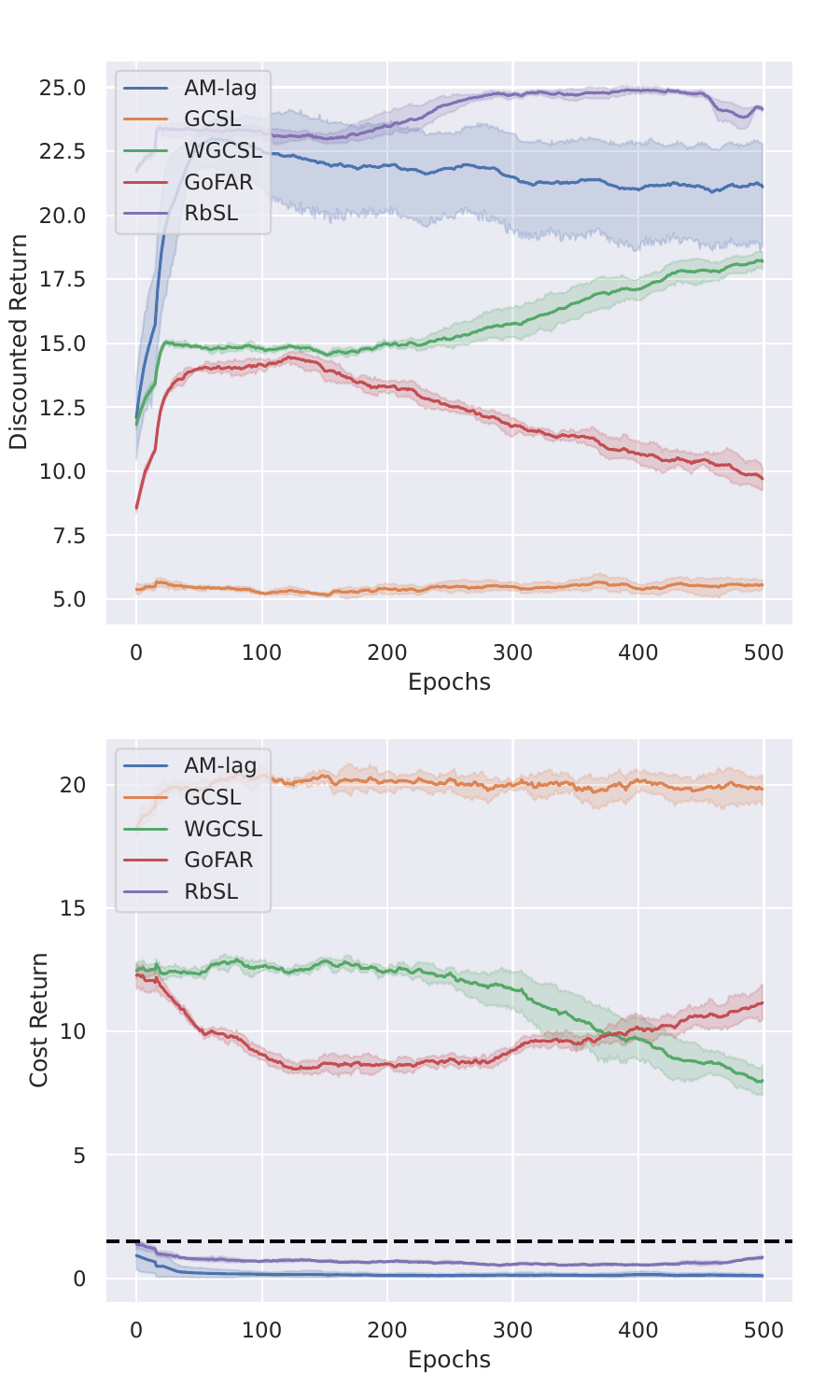}\\
    \end{minipage}%
}%
\subfigure[PickAndPlace]{
    \begin{minipage}[t]{0.2\pdfpagewidth}
        \centering
        \includegraphics[width=\linewidth]{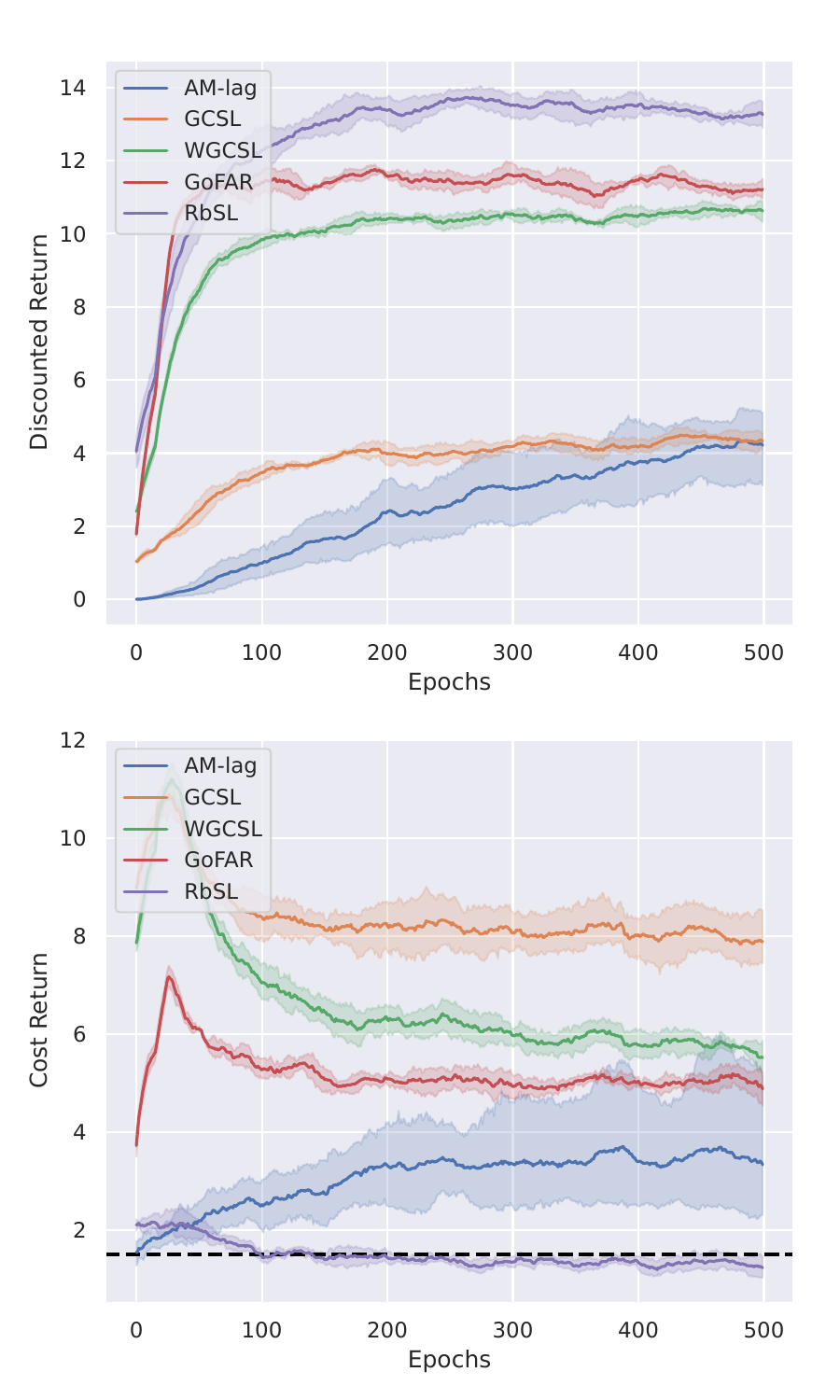}\\
    \end{minipage}%
}%
\subfigure[Push]{
    \begin{minipage}[t]{0.2\pdfpagewidth}
        \centering
        \includegraphics[width=\linewidth]{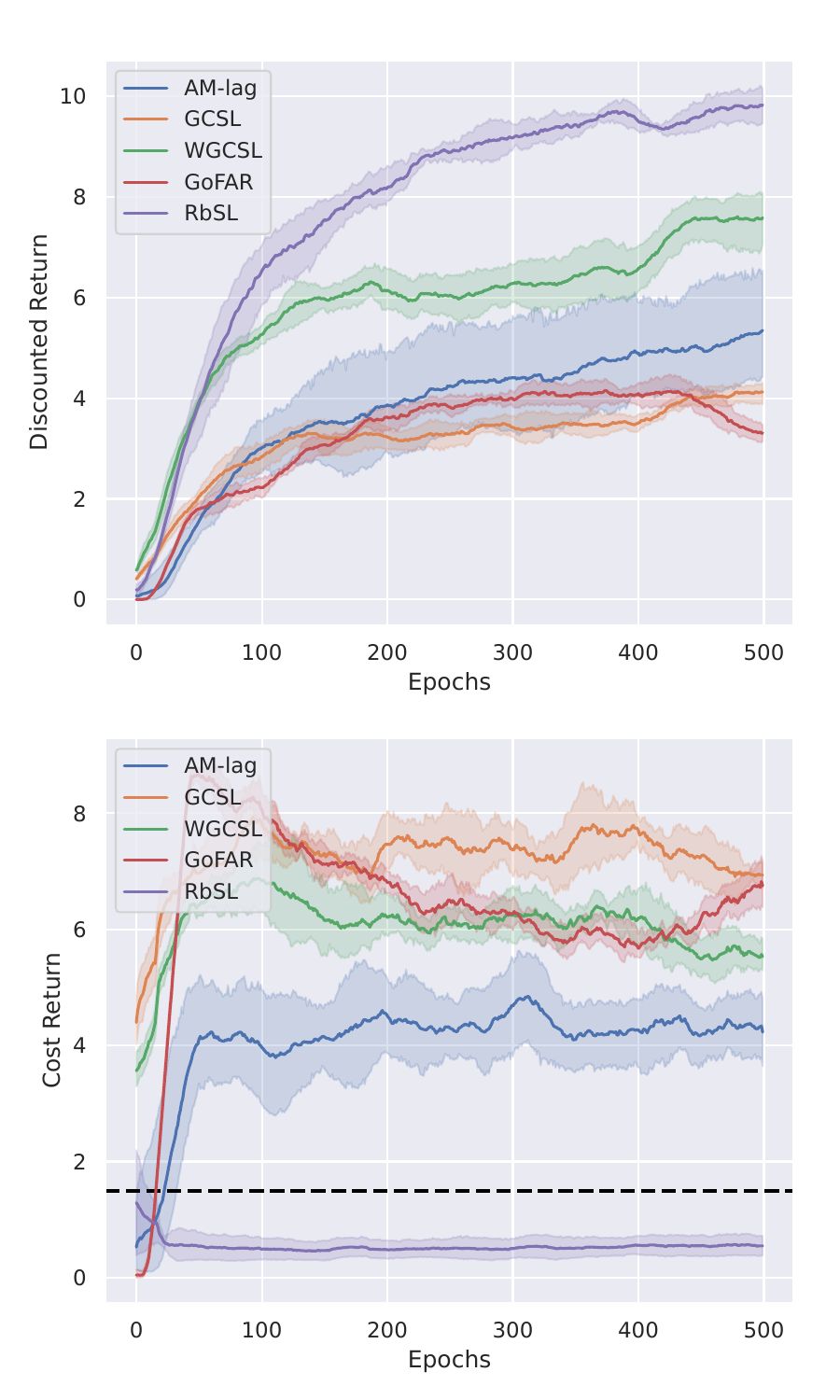}\\
    \end{minipage}%
}%
\subfigure[Slide]{
    \begin{minipage}[t]{0.2\pdfpagewidth}
        \centering
        \includegraphics[width=\linewidth]{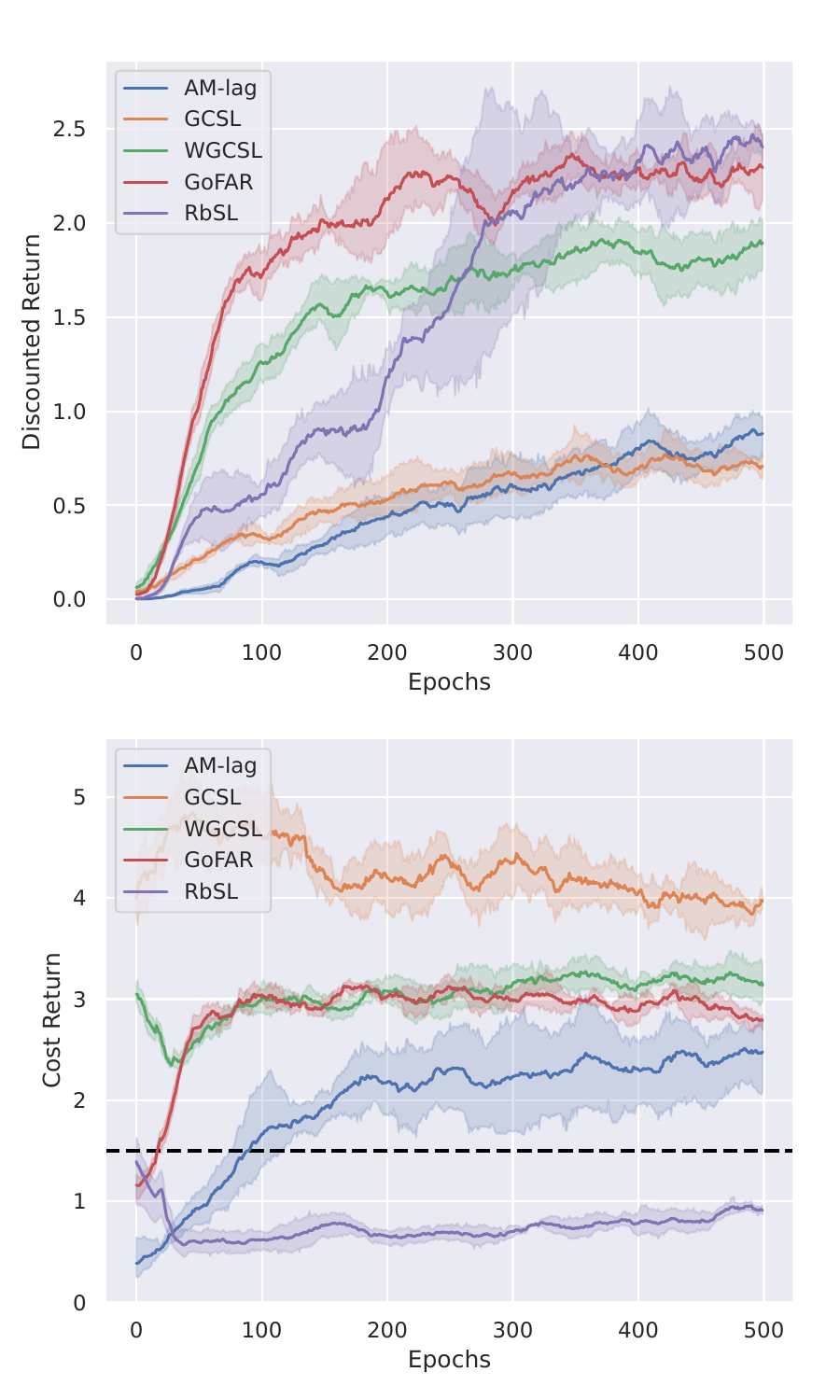}\\
    \end{minipage}%
}%
\centering
\caption{Training curves of \textit{PushObstacle, PickAndPlaceObstacle, SlideObstacle} in the 0.5-0.5 setting and \textit{ReachObstacle} in the 0-1 setting. In each figure, the sub-figures show the average number of discounted returns and cost returns per epoch, where the black dotted line shows the safe constraint limit. }
\label{fig:curves}
\end{figure*}

In all tasks, the rewards are sparse and binary: the agent receives +1 if it achieves a desired goal and 0 otherwise. We collect data with two different policies for offline training, like the way in \cite{yang2021rethinking}. The random dataset is collected by a uniform random policy. And the expert chooses an online trained data-augment TD3+HER policy \cite{lin2020invariant, fujimoto2018addressing} with Gaussian noise. Each dataset has $2\times 10^6$ transitions, enough for offline algorithms to learn a qualified policy. In our experiments, we mix the random data and expert data in different proportions to test the policy's robustness. The main results are presented in Table \ref{tab:main} and Figure \ref{fig:curves}. 

\begin{figure}[htpb]
\centering
\subfigure[Gym-Robotics]{
    \begin{minipage}[t]{0.125\pdfpagewidth}
        \centering
        \includegraphics[height=2.5cm, width=2.5cm]{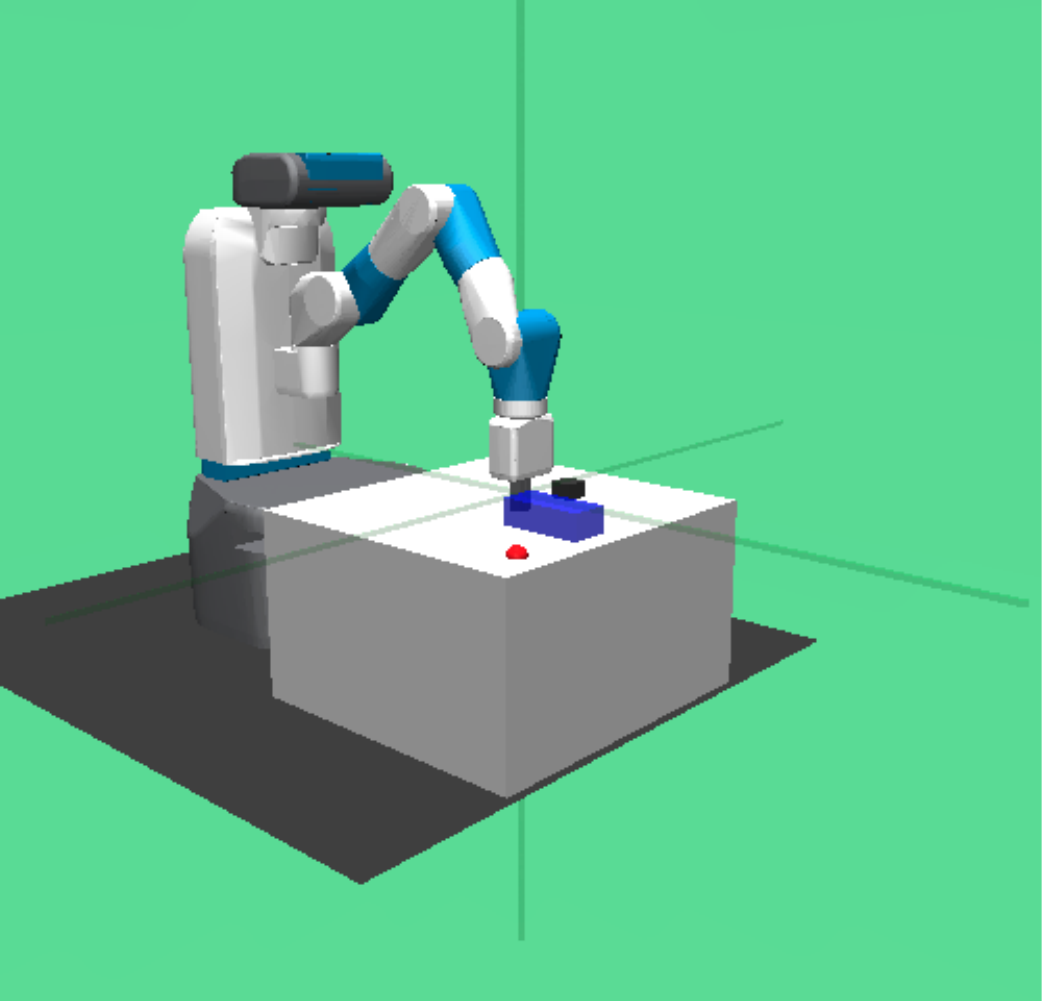}\\
    \end{minipage}%
}
\subfigure[Panda-Gym]{
    \begin{minipage}[t]{0.125\pdfpagewidth}
        \centering
        \includegraphics[height=2.5cm, width=2.5cm]{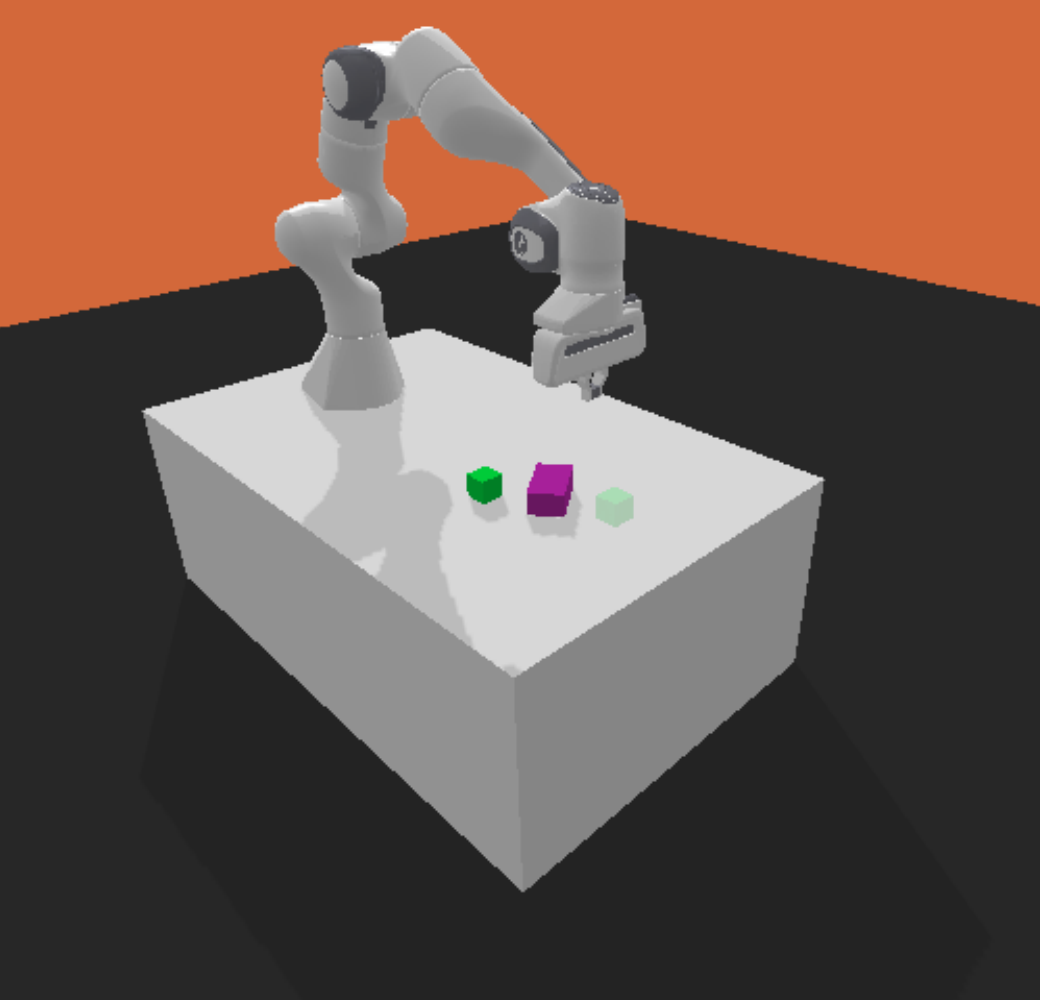}\\
    \end{minipage}%
}%
\subfigure[Franka Emika Panda]{
    \begin{minipage}[t]{0.125\pdfpagewidth}
        \centering
        \includegraphics[height=2.5cm, width=2.5cm]{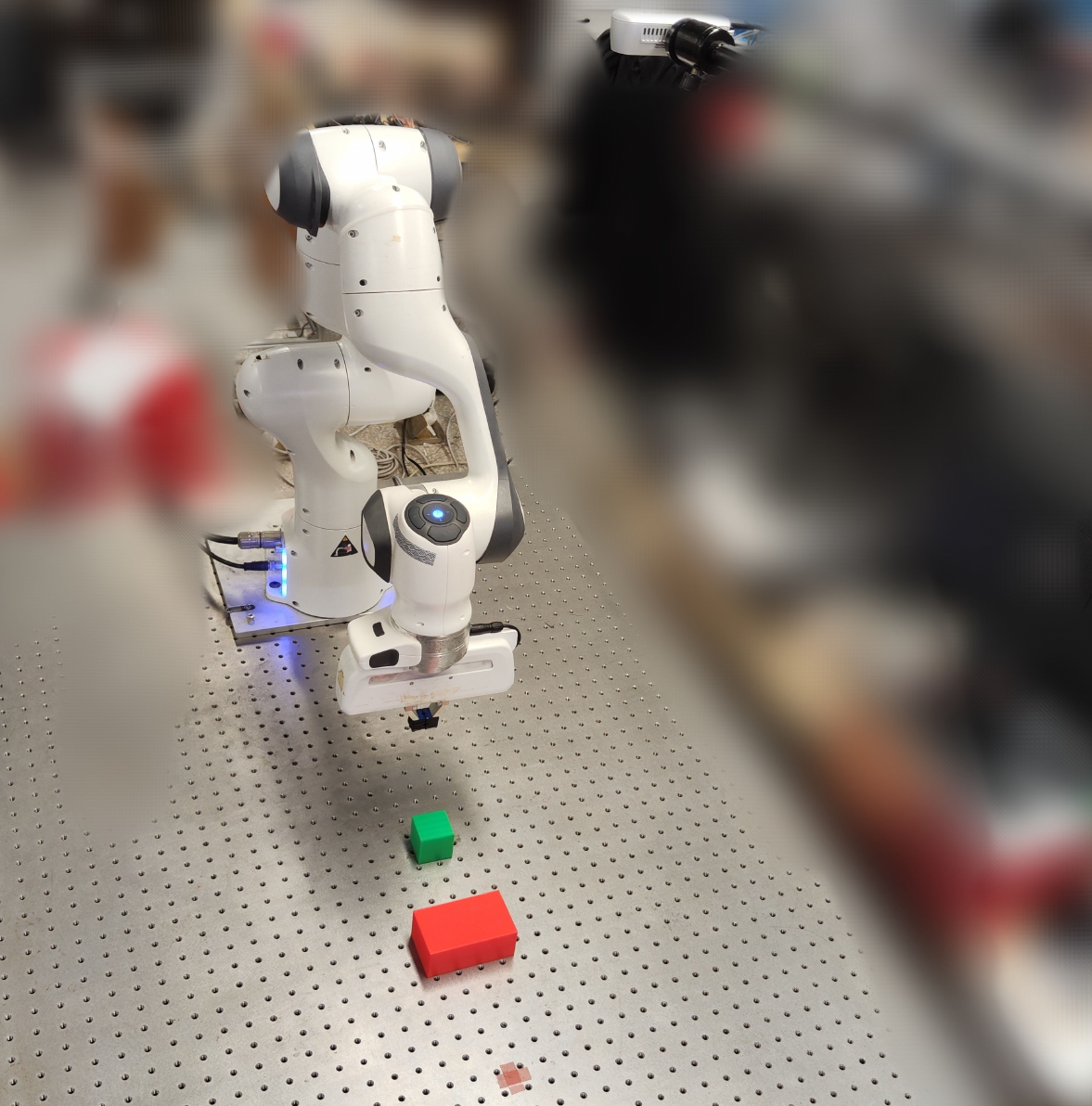}\\
    \end{minipage}%
}%
\centering
\caption{Goal-conditioned environments: (a) MuJoCo Gym-Robotics, (b) Panda-Gym, a simulation environment for the real-world experiment, (c) An experiment environment for the Franka Emika Panda robotic arm.}
\label{fig:envs}
\end{figure}

For the real-world experiment, we employ a Franka Emika Panda robot for manipulation and an Intel RealSense camera which is set beyond the robotic arm to capture object coordinates. Its setup is visualized in Figure \ref{fig:envs}(c).

\subsection{Baselines}

We compare our method with prior offline GCRL and offline safe RL algorithms. {GCSL}\cite{ghosh2020learning} is an imitation learning method which shows good performance in goal-reaching tasks. {WGCSL}\cite{yang2021rethinking} solves the offline GCRL problem by improving GCSL in the offline setting using AWR\cite{peng2019advantage}. {GoFAR}\cite{ma2022offline} is a state-of-the-art offline GCRL method that is practical in various challenging problem settings. Since GCRL methods can learn policies to reach goals, they may cause collisions with obstacles. Therefore, we use {AM-lag}\cite{chebotar2021actionable}, an offline safe GCRL, to compare the safety with our algorithm. It is built on a conservative Q-Learning\cite{kumar2020conservative} method using goal-chaining and simply adds a Lagrangian multiplier to penalize constraint violations. 

\subsection{Evaluations and Results}

We evaluate our method in four tasks and record the success rate and cost return in Table \ref{tab:main}. The constraint limit is set as 1.5 in all tasks, which means the average constraint violation rate is 1.5 every episode. The Lagrangian approach uses adaptive PID-based Lagrangian multipliers\cite{stooke2020responsive} and the initial value of the lagrangian multiplier is set as 10 which should be set as large as possible. As we can see, it becomes evident that RbSL consistently performs well across all settings, achieving the highest average success rate and the lowest average cost return. In other words, RbSL can reach goals precisely without encountering constraint violations. However, we observed that our method underperforms compared to GoFAR in \textit{SlideObstacle}, primarily because the recovery policy faces challenges when handling complex tasks that are less safety-critical. The previous GCRL methods like GoFAR often lead to more collisions with obstacles, even though they exhibit good task performance. Conversely, offline safe RL methods encounter difficulties in increasing the success rate when attempting to violate the constraints. Nevertheless, RbSL consistently demonstrates robust and outstanding performance across datasets of various quality. To better visualize the effects of the algorithm, we show training curves in Figure \ref{fig:curves}. As evident, there is a significant gap between RbSL and the baselines, especially the cost return in challenging tasks like \textit{PushObstacle, SlideObstacle}. Besides, RbSL exhibits faster convergence compared to the others. Overall, RbSL achieves superior task performance by effectively avoiding obstacles, thereby reducing the likelihood of constraint violations. 


\subsection{Real World Deployment}

To apply our algorithm to a real robot, we built a simulation environment in panda-gym\cite{gallouedec2021panda}. The agent receives joint positions and velocities of the end-effector as a robotic state and it is controlled by a 3-dim vector describing the end-effector’s position in simulation. We compare RbSL with WGCSL for 100 epochs of training. The simulation result is shown in Table \ref{tab:sim}.

Finally, we apply the offline trained model on the real manipulator and use joint position control without sim-to-real fine-tuning. We use HSV color space filtering to obtain the positions of objects and obstacles in the real world. The task is pushing the green object to the goal without colliding with the red obstacle. The results are shown in Figure \ref{fig:real} and more details can be found in the supplementary video.

\begin{figure}[t]
\centering
\includegraphics[trim=0cm -0cm 0cm 0cm,clip,width=\linewidth]{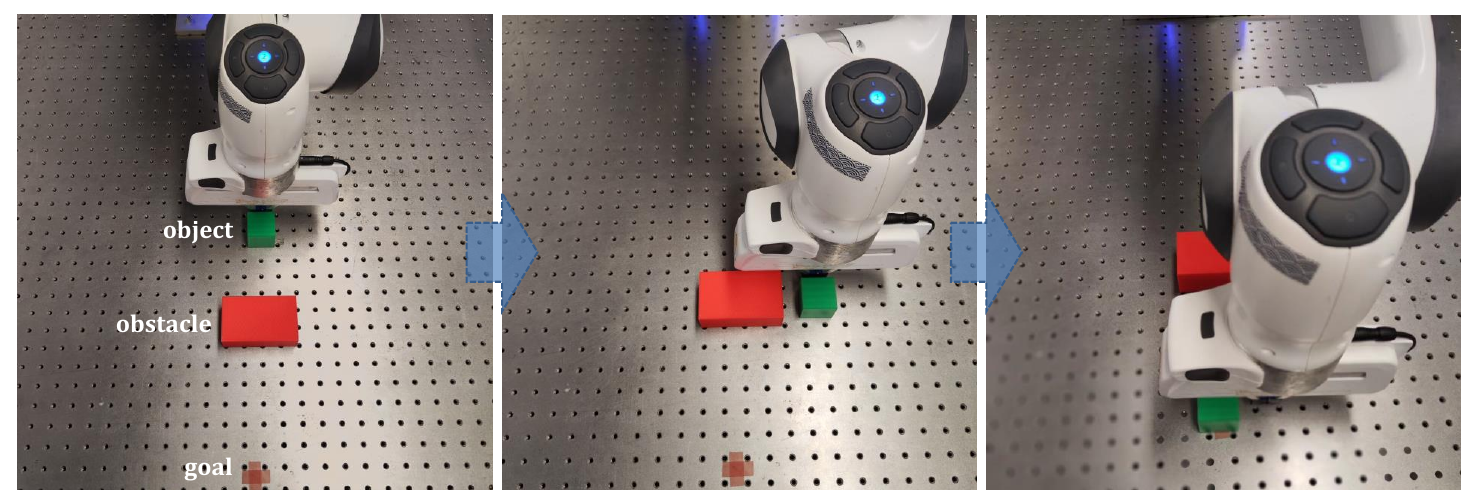}\\
\centering
\caption{Leveraging RbSL, we display the example execution trajectory of push, which involves navigating around an obstacle to reach the goal.}
\label{fig:real}
\end{figure}

\renewcommand{\arraystretch}{1.1}
\begin{table}[t]
  \centering
    \caption{Simulation Result}
    \resizebox{\linewidth}{!}{\begin{tabular}{cccc}
    \toprule
    \textbf{Algorithm} & \textbf{Success Rate} & \textbf{Discounted Return} & \textbf{Cost Return} \\
    \midrule
    WGCSL\cite{yang2021rethinking} & $78.3\%$ & $11.862$ & $1.493$ \\
    \textbf{RbSL(Ours)} & $\textbf{96.7\%}$ & $\textbf{17.056}$ & $\textbf{0.978}$ \\
    \bottomrule
  \end{tabular}}
  \label{tab:sim}
\end{table}

\section{CONCLUSIONS}

This paper proposes a novel offline GCRL method to solve constraints in complex environments by combining a goal-conditioned policy and a recovery policy. In our experiment, it is proved that the proposed RbSL can accomplish goal-reaching tasks with lower cost return and it outperforms the prior methods. Additionally, RbSL is tested in a real robot and achieves accomplished returns.





\section*{ACKNOWLEDGMENT}

This work was supported by the National Natural Science Foundation of China (No.62103225) and the Natural Science Foundation of Shenzhen (No.JCYJ20230807111604008).


\bibliographystyle{ieeetr}
\bibliography{reference}

\end{document}